\documentclass[letterpaper]{article}
\usepackage[preprint]{aaai2027}
\usepackage[hyphens]{url}
\usepackage{graphicx}
\usepackage{natbib}
\usepackage{caption}
\usepackage{booktabs}
\usepackage{amsmath}
\usepackage{amssymb}
\usepackage{algorithm}
\usepackage{algorithmic}
\usepackage[table]{xcolor}
\usepackage{tabularx}

\newcommand{\csca}{\textsc{Maga}\xspace}
\usepackage{xspace}

\title{MAGA: \underline{M}ulti-Pl\underline{a}tform Self-Fusion of \underline{G}UI \underline{A}gents\\via Structured Action Distillation}
\author{Hang Yan\textsuperscript{1,2}\thanks{Work done during internship at Ant Group.} \quad
Zhangxuan Gu\textsuperscript{2}\corresponding \quad
Beitong Zhou\textsuperscript{2}\quad
Jiaxuan Chen\textsuperscript{2,3} \quad
Runze Li\textsuperscript{2}\quad
Yusong Hu\textsuperscript{2} \\
Shuheng Shen\textsuperscript{2}\quad
Changhua Meng\textsuperscript{2}}
\affiliations{
\textsuperscript{1}Xi'an Jiaotong University
\quad \textsuperscript{2}Ant Group \quad \textsuperscript{3}Shanghai Jiao Tong University\\
hyan@stu.xjtu.edu.cn \quad guzhangxuan.gzx@antgroup.com
}
\begin{document}
\maketitle

\begin{abstract}
Graphical user interface (GUI) agents based on large language models are increasingly deployed across mobile, web, and desktop environments. However, existing agents are typically domain-specific, limiting the deployment and user experience. This motivates the consolidation of specialized models into a single cross-environment policy. 
Weight merging directly merges domain-specific experts but can corrupt executable actions under expert disagreement, while on-policy distillation (OPD) avoids conflicting teacher supervision yet still treats all response tokens equally during distillation, ignoring that action tokens are the only interface between the environment and the agent. 
To address this, We introduce \csca that re-allocates training signal according to the structured action. 
Based on the correctness of the generated action, it suppresses unnecessary or invalid distillation signals and focuses learning on erroneous actions. 
Besides, a training-only hint optimizes the supervision signal provided by domain-specific teachers without changing the student input.
Across two model scales, \csca achieves the highest mean success rate, outperforming the strongest baseline by $2.0\%$ at 8B and achieves almost the same average performance with teachers.
\end{abstract}

\section{Introduction}

\begin{figure}[t]
\centering
\includegraphics[width=\columnwidth]{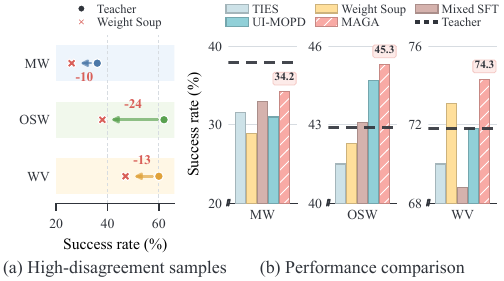}
\caption{Results from three benchmarks, including MobileWorld (MW), OSWorld (OSW), and WebVoyager (WV).
(a) On 900 random samples, we identify 66 tasks where domain-specific models exhibited high disagreement, and weight merging reduces
success rate of the merged model by 10\%--24\%.
(b) Success rate of models based on Qwen3-VL-8B across three benchmarks. The horizontal dashed lines mark the
domain-specific teachers. \csca is the
strongest unified model on every benchmark and exceeds the corresponding
domain-specific teacher on OSWorld and WebVoyager.}
\label{fig:token-light-action}
\end{figure}

Recent advances in vision-language models (VLMs)~\citep{zhu2024minigpt,li2025survey,zhang2026efficient,liang2026comprehensive,zhang2026cofft} have improved visual
understanding, multimodal reasoning, and grounded decision making.
These capabilities support graphical user interface (GUI) agents~\citep{zhou2025mai,xu2026mobile,team2026ui,tang2026gui2} that interpret
screen content and execute actions from natural language instructions.
Earlier GUI agents typically focused on mobile applications~\citep{sun2025genesis,cheng2026openmobile,gong2026venusbench}, web browsers~\citep{yao2022webshop,li2026online,zhang2026webarbiter}, or desktop
operating systems~\citep{liu2025scalecua,jia2025agentstore,xue2026evocua} separately.  These approaches introduce deployment complexity and limit cross-domain user experience, so recent systems combine several of these domains
into a single agent.

Previous methods combine abilities across domains either by merging expert
parameters.
Weight merging combines separately trained domain-specific experts, as in
Model Soup~\citep{wortsman2022model} and TIES~\citep{yadav2023ties}.
GUI domains share action types such as \texttt{Click} and \texttt{Scroll}, but
their domain-specific models can disagree on the corresponding decisions.
These disagreements can degrade the merged model.
As shown in Figure~\ref{fig:token-light-action}(a), when evaluated on samples with high disagreement among domain-specific models, the action success rate of the merged model drops by $10\%$--$24\%$ compared to the individual models. 

Alternatively, per-sample routed on-policy distillation (OPD)~\citep{lu2025onpolicydistillation,xiao2026mimo,yang2026opidonpolicyskilldistillation,wu2026seedselfevolvingonpolicydistillation} transfers each expert's supervision to the student without combining their parameters.
For each student-generated response, only the corresponding frozen
domain-specific model, which serves as the teacher, scores the sampled tokens.
However, ordinary OPD treats the response as a flat sequence and assigns equal
weight to all tokens.
This allocation is poorly matched to GUI scenarios, where the final short structured action is the only part executed by the agent. 
Moreover, GUI actions directly affect the interface state, and incorrect actions can significantly shift the system away from its intended state.

To address this, we calibrate the  detailed distillation signal during training for GUI actions. A GUI
action comprises an action type and its associated parameters(if required).
Uniformly increasing supervision across all action tokens ignores this
structure.
Consequently, we introduce \csca, a distillation method that allocates additional training supervision according to the structure of the action.
\textbf{It operates on both the student and teacher sides.}
\textbf{On the student side}, \csca removes fully correct responses from
distillation and focuses extra supervision on erroneous action components.
It amplifies the full action span when the action type is correct. Otherwise,
it amplifies only the type and masks the incompatible parameters.
\textbf{On the teacher side}, we condition the domain-specific teacher on a
hint of the correct action during training. The hint
therefore provides a more reliable signal when the teacher scores the
student-sampled tokens.
The student never receives this hint, so its input and output remains unchanged.

We evaluate \csca on MobileWorld, OSWorld, and WebVoyager at two model scales.
As shown in Figure~\ref{fig:token-light-action}(b), at 8B it achieves a mean
success rate of $51.2\%$, exceeding the strongest baseline by $2.0\%$ and achieves almost the same average performance with teachers.
Our contributions can be concluded as follows:

\begin{itemize}
    \item We identify two limitations of existing approaches to unifying
    domain-specific GUI agents: weight merging degrades when the experts
    disagree, and ordinary OPD under-allocates signal to short structured actions.
    \item We introduce \csca, which re-allocates distillation signal according to
    the action structure through student-side conditional training signal re-allocation and a training-only
    teacher hint.

    \item Across three GUI domains and two model scales, \csca achieves the
    highest mean SR among unified methods. At 8B, it
    exceeds the strongest baseline by $2.0\%$ and achieves almost the same average performance with teachers.
\end{itemize}


\section{Related Work}

\paragraph{GUI agents.}
GUI agents now operate across increasingly diverse interfaces and tasks. Browser
agents~\citep{yao2022webshop,zheng2024gpt,zhang2026webarbiter,li2026online}
perceive and act on changing websites. Moving to
mobile devices~\citep{sun2025genesis,tang2026gui,gong2026venusbench,cheng2026openmobile}
introduces persistent app state, cross-app dependencies, and longer action
sequences. Desktop control
agents~\citep{jia2025agentstore,liu2025scalecua,yang2026symphony,xue2026evocua} operate over
applications, files, menus, and system tools. Despite these interface
differences, practical deployment benefits from generalist
agents~\citep{cheng2024seeclick,zhou2025venusbench,gu2025ui,wu2025atlas,hu2026gui,team2026ui,xu2026mobile}
that cover several interface families within one model. Following this line of
work, we study how to combine separately trained per-domain GUI models into a
single agent.

\paragraph{Model merging.}
Training a single model on mixed-domain trajectories can improve one domain at
the expense of another~\citep{xiao2026mimo,zhang2026beyond,xu2026deepseek}, including under
supervised fine-tuning (SFT)~\citep{luong2024reft} or reinforcement learning
(RL)~\citep{shao2024deepseekmath}. Post-hoc merging instead combines
specialized checkpoints, as in Weight Soup~\citep{wortsman2022model} and
TIES~\citep{yadav2023ties}. For GUI agents, domain-specific teachers share a
structured action space of action types and optional parameters. Merging is
largely harmless when they agree, but parameter disagreements can shift the
prediction away from the corresponding domain-specific teacher
(Figure~\ref{fig:token-light-action}(a)). OPD~\citep{wu2026seed,yang2026opid,xiao2026mimo,xu2026deepseek}
avoids this conflict by scoring each student-generated token with that teacher.
However, existing OPD method~\citep{lian2026uimopd} distributes the original
token-level signal across long reasoning traces and a few action tokens, leaving
the latter under-supervised. Existing evaluation also covers only two domains,
leaving broader scalability untested. We therefore introduce \csca to allocate
signal according to the structured action and generalize it to three GUI
domains.

\begin{figure*}[t]
\centering
\includegraphics[width=\textwidth]{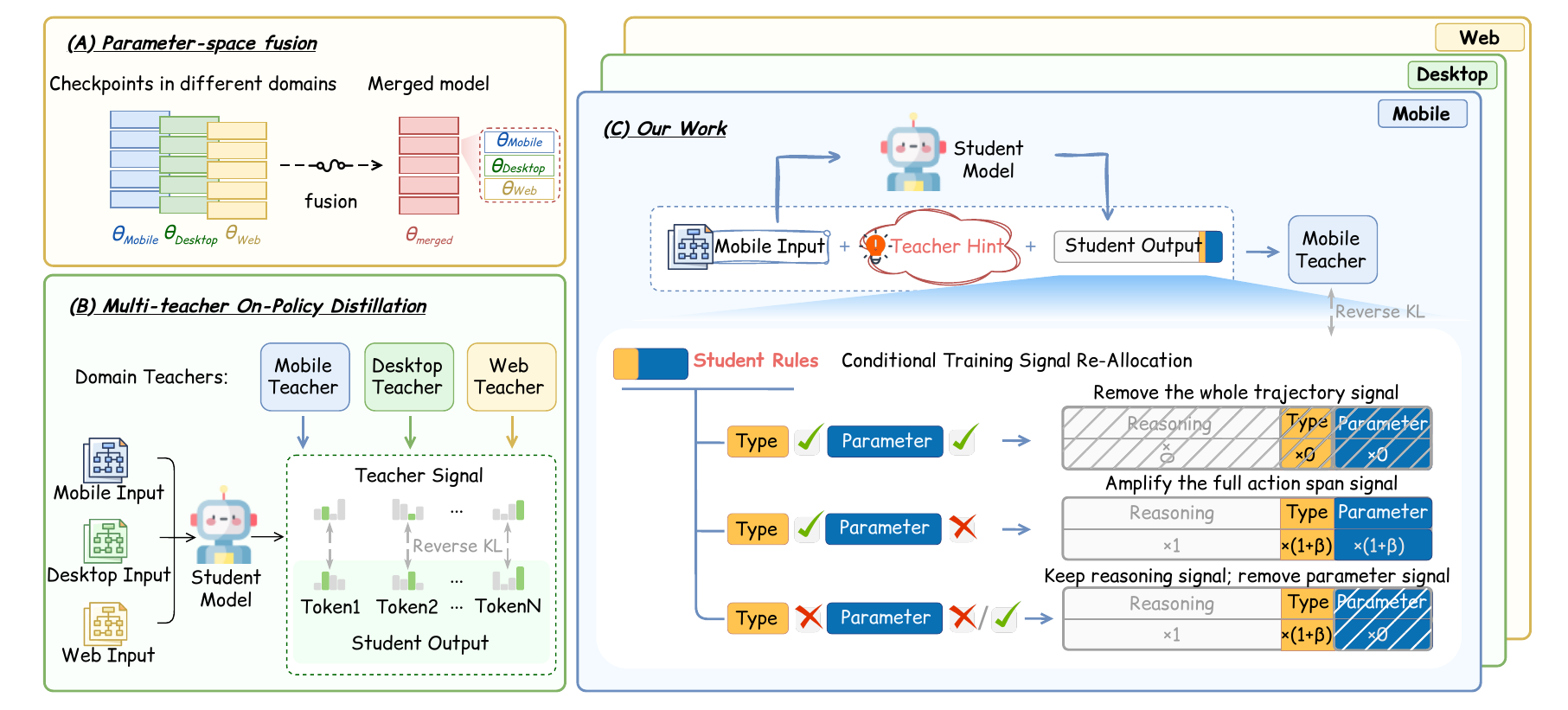}
\caption{Comparison of three strategies for unifying domain-specific GUI agents.
(a) Parameter merging combines the weights of domain-specific teachers, which can alter executable decisions when the teachers disagree.
(b) Per-sample routed on-policy distillation (OPD) trains a student through token-level teacher supervision, but still treats the response as a flat token sequence, leaving the action tokens under-supervised.
(c) \csca allocates distillation signal according to the action grammar. On the student side, it removes fully correct responses from distillation. For an incorrect response with the correct action type, it amplifies the full action span. When the type is wrong, it amplifies only the type token and masks the downstream parameters. On the teacher side, a training-only hint improves supervision signal without changing the student input.}
\label{fig:method-overview}
\end{figure*}

\section{Method}
\label{sec:method}

This section first introduces the GUI agent interface and routed OPD
(Figure~\ref{fig:method-overview}(b)). We then present \csca
(Figure~\ref{fig:method-overview}(c)), which combines
\textbf{student-side conditional signal allocation}
(Section~\ref{sec:student-side-allocation}) with a
\textbf{teacher-side hint} (Section~\ref{sec:teacher-side-hint}) to provide
targeted supervision for short, structured action outputs.

\subsection{Preliminaries}
\label{sec:prelim}

\subsubsection{GUI Agent Interface}

During each interaction step, a GUI agent receives a task instruction $u$, the current visual observation
$o$, and an optional interaction history $H$. We write the resulting model
input as $x=(u,o,H)$. For each model call, the agent produces a response
$y=(r,a)$ containing reasoning $r$ and one executable action $a$.

The action is structured as $a=(z,p_z)$, where $z$ is a discrete action type
and $p_z$ contains the parameters required by that type. The action type $z$ selects the
parameter schema $p_z$, which may be empty. These parameters may
be screen coordinates, text, a URL, or a key combination. Coordinates are
normalized to the shared $[0,1000]$ interface, which are one parameter class
used by spatial actions.

\subsubsection{Routed On-Policy Distillation}


Let $d$ denote a GUI domain, $x\sim\mathcal{D}_d$ an input, and $T_d$ a frozen domain-specific teacher. For a student $\pi_\theta$ generating tokens $y=(y_1,\ldots,y_{|y|})$ with prefix $y_{<t}$, calculating the exact reverse KL divergence against $T_d$ requires summing over the full vocabulary. For efficiency, we approximate this using only the single sampled token $y_t$ following~\citep{li2026rethinking}. We define the token-level distillation advantage as the gradient-stopped difference in log probabilities:
\begin{equation}
\widehat{A}_t^{\mathrm{OPD}} =
\operatorname{sg}\!\left[
\log \pi_{T_d}(y_t\mid x,y_{<t})
-\log \pi_\theta(y_t\mid x,y_{<t})
\right].
\label{eq:distill-adv}
\end{equation}
Routed OPD then optimizes the student using this advantage-weighted objective:
\begin{equation}
\mathcal{L}_{\mathrm{OPD}}(\theta)
=-\mathbb{E}_{d,x,y}\Bigg[ 
\frac{1}{|y|}\sum_{t=1}^{|y|}
\widehat{A}_t^{\mathrm{OPD}}
\log\pi_\theta(y_t\mid x,y_{<t})
\Bigg].
\label{eq:opd}
\end{equation}

\subsection{Student-Side Conditional Training Signal Re-allocation}
\label{sec:student-side-allocation}
A GUI action is represented as $a=(z,p_z)$, where the action type $z$ selects a
type-specific parameter schema $p_z$. For example, coordinates, text, URLs, and key
combinations are different parameter classes within this structure. Let
$\mathcal{I}_{\mathrm{reason}}$, $\mathcal{I}_{\mathrm{type}}$, and
$\mathcal{I}_{\mathrm{param}}$ denote the reasoning, action-type, and
parameter token positions, respectively.
This decomposition yields three training cases, and we use a token weight $w_t$
to scale the routed-OPD advantage at token position $t$.

{\textbf{Training Rule 1: Correct action type, correct parameters.}}
When both action type $z$ and parameter $p_z$ are completely correct, the rollout receives the
maximum reward. We remove the whole trajectory from distillation by setting its
advantage to zero:
\[
w_t=
\begin{cases}
0, & t\in\mathcal{I}_{\mathrm{reason}},\\
0, & t\in\mathcal{I}_{\mathrm{type}},\\
0, & t\in\mathcal{I}_{\mathrm{param}}.
\end{cases}
\]
Thus, reasoning, action-type, and parameter tokens receive no distillation
signal. The filter avoids further optimization of an action that the evaluator
already accepts completely.

{\textbf{Training Rule 2: Correct action type, incorrect parameters.}} Unlike a
fully successful rollout, this sample contains a remaining error that the
teacher can correct. We assign
\[
w_t=
\begin{cases}
1, & t\in\mathcal{I}_{\mathrm{reason}},\\
1+\beta, & t\in\mathcal{I}_{\mathrm{type}},\\
1+\beta, & t\in\mathcal{I}_{\mathrm{param}}.
\end{cases}
\]
Reasoning tokens retain their ordinary routed-OPD weight, while the complete
action span, including both type and parameters, is amplified by
$1+\beta$. This prevents the training signal of the action tokens from being
diluted by the longer response.

{\textbf{Training Rule 3:    Incorrect action type (regardless of parameters).}}
When the action type is incorrect, the discrete type decision is the primary
error to correct. Moreover, the predicted type selects the wrong parameter
schema, so its downstream parameters are not semantically valid supervision.
We therefore amplify the type tokens and mask the parameter tokens.
For a rollout with an incorrect action type, we assign token weight $w_t$ as follows:
\[
w_t=
\begin{cases}
1, & t\in\mathcal{I}_{\mathrm{reason}},\\
1+\beta, & t\in\mathcal{I}_{\mathrm{type}},\\
0, & t\in\mathcal{I}_{\mathrm{param}}.
\end{cases}
\]
Thus, reasoning tokens retain their original signal, the incorrect action type
receives additional correction, and downstream parameters receive no
distillation signal.

\subsection{Teacher-Side Hint}
\label{sec:teacher-side-hint}
On the teacher side, we modify only the input prompt of the routed
domain-specific teacher. Let $\mathcal{P}(x)$ denote the standard prompt
constructed from the complete input $x$, including the system instruction,
interaction history, screenshot, and current task. Let $h(z^*)$ be a hint
containing only the reference action type, and let $\oplus$ append this hint
to the end of the final user message. The teacher and student prompts are
\[
\begin{aligned}
\mathcal{P}_T(x,z^*)
&=\mathcal{P}\!\left(x\oplus h(z^*)\right),\\
\mathcal{P}_S(x)
&=\mathcal{P}(x).
\end{aligned}
\]
Thus, the teacher prompt differs from the student prompt only by the appended
hint. All other input content remains identical. The hint contains no
coordinates, text, URLs, keys, or reasoning and is never added to the student
prompt.

The student rollout remains unchanged. The teacher does not decode a response
and only scores the tokens sampled by the student. This changes the teacher
term in the distillation advantage but does not alter the student-side weight
$w_t$ assigned above. The updated advantage for token $y_t$ is
\begin{equation}
\begin{aligned}
\widehat{A}_t^{\mathrm{hint}}
&=\operatorname{sg}\!\Big[
\log \pi_{T_d}\!\left(y_t\mid\mathcal{P}_T(x,z^*),y_{<t}\right)\\
&\qquad
-\log \pi_\theta\!\left(y_t\mid\mathcal{P}_S(x),y_{<t}\right)
\Big].
\end{aligned}
\label{eq:hint-adv}
\end{equation}

\subsection{Overall Framework}
\label{sec:unified-objective}

Figure~\ref{fig:method-overview}(c) summarizes the complete framework. For
each offline input, the student samples a one-step response and is assigned to one of the three allocation cases in
Section~\ref{sec:student-side-allocation}. The corresponding token weights
$w_t$ determine which parts of the routed OPD signal are retained, amplified,
or masked. Meanwhile, the routed domain-specific teacher scores the same
student-sampled tokens using the type-hinted prompt in
Section~\ref{sec:teacher-side-hint}. Combining the student-side weights with
the type-hinted teacher advantage gives the final objective:
\begin{equation}
\begin{aligned}
\mathcal{L}_{\csca}(\theta)
&=-\mathbb{E}_{d,x,y}\Bigg[ \\
&\quad \frac{1}{|y|}\sum_{t=1}^{|y|}
w_t\widehat{A}_t^{\mathrm{hint}}
\log\pi_\theta(y_t\mid x,y_{<t})
\Bigg].
\end{aligned}
\label{eq:csca-loss}
\end{equation}

During training, all domain-specific teachers remain frozen, and only the student is updated. 
When both student-side allocation and teacher-side hinting are disabled, the objective reduces to ordinary routed OPD.
Detailed training algorithm is provided in
Appendix~\ref{sec:hint-allocation-implementation}.

\begin{table*}[t]
\centering
\footnotesize
\setlength{\tabcolsep}{2.5pt}
\begin{tabularx}{0.98\textwidth}{@{}l*{8}{>{\centering\arraybackslash}X}@{}}
\toprule
Method & \multicolumn{2}{c}{MobileWorld} & \multicolumn{2}{c}{OSWorld} &
         \multicolumn{2}{c}{WebVoyager} & Mean SR $\uparrow$ & TNS (\%) $\uparrow$ \\
\cmidrule(lr){2-3}\cmidrule(lr){4-5}\cmidrule(lr){6-7}
 & SR (\%) $\uparrow$ & $\Delta T$ & SR (\%) $\uparrow$ & $\Delta T$
 & SR (\%) $\uparrow$ & $\Delta T$ & & \\
\midrule
\multicolumn{9}{c}{\textbf{Qwen3-VL-2B}} \\
\addlinespace[2pt]
\rowcolor{gray!12}
Domain-specific teacher     & 24.8 & - & 25.2 & - & 47.9 & - & 32.6 & 100.0 \\
Student                     & 3.4 & -21.4 & 7.9 & -17.3 & 7.9 & -40.0 & 6.4 & 20.5 \\
SFT                         & 11.1 & -13.7 & 20.6 & -4.6 & 33.6 & -14.3 & 21.8 & 65.6 \\
GRPO                        & 6.8 & -17.9 & 11.1 & -14.1 & 27.9 & -20.0 & 15.3 & 43.3 \\
Weight Soup                 & \underline{12.0} & \underline{-12.8} & \underline{21.7} & \underline{-3.5} & \textbf{40.0} & \textbf{-7.9} & \underline{24.5} & \underline{72.6} \\
TIES                        & \textbf{14.5} & \textbf{-10.3} & 20.6 & -4.6 & 34.3 & -13.6 & 23.1 & 70.7 \\
UI-MOPD                     & \underline{12.0} & \underline{-12.8} & 21.4 & -3.8 & 30.0 & -17.9 & 21.1 & 65.3 \\
\midrule
\csca (ours)                & \textbf{14.5} & \textbf{-10.3} & \textbf{23.6} & \textbf{-1.6} & \underline{37.9} & \underline{-10.0} & \textbf{25.3} & \textbf{77.1} \\
\midrule
\multicolumn{9}{c}{\textbf{Qwen3-VL-8B}} \\
\addlinespace[2pt]
\rowcolor{gray!12}
Domain-specific teacher     & 37.6 & - & 42.8 & - & 72.1 & - & 50.9 & 100.0 \\
Student                     & 12.8 & -24.8 & 20.9 & -22.0 & 37.1 & -35.0 & 23.6 & 44.8 \\
SFT                         & \underline{33.3} & \underline{-4.3} & 43.1 & +0.3 & 68.6 & -3.6 & 48.3 & 94.8 \\
GRPO                        & 17.9 & -19.7 & 25.5 & -17.3 & 60.0 & -12.1 & 34.5 & 63.5 \\
Weight Soup                 & 29.1 & -8.5 & 42.3 & -0.5 & \underline{72.9} & \underline{+0.7} & 48.1 & 92.3 \\
TIES                        & 31.6 & -6.0 & 41.5 & -1.4 & 70.0 & -2.1 & 47.7 & 92.7 \\
UI-MOPD                     & 30.8 & -6.8 & \underline{44.7} & \underline{+1.9} & 72.1 & 0.0 & \underline{49.2} & \underline{95.4} \\
\midrule
\csca (ours)                & \textbf{34.2} & \textbf{-3.4} & \textbf{45.3} & \textbf{+2.4} & \textbf{74.3} & \textbf{+2.1} & \textbf{51.2} & \textbf{99.9} \\
\bottomrule
\end{tabularx}
\caption{Main comparison on MobileWorld, OSWorld, and WebVoyager
across two model scales. For each domain, $\Delta T$ is the method SR minus its
domain-specific teacher SR. SRs are derived from integer success counts over
117, 369, and 140 tasks, respectively. TNS denotes the Teacher-Normalized
Score defined in Eq.~\ref{eq:tns}. Among unified methods within each model
scale, the best results are in \textbf{bold} and the second-best results are
\underline{underlined}. Uparrow indicates that a higher value is better}
\label{tab:main_results}
\end{table*}

\section{Experiments}
In this section, we first describe the experiment implementation details, followed by presenting the main results of our method compared to various baselines and an ablation study. Then we provide deeper analysis of the trained student's performance guided by six key questions, revealing fine-grained behavior beyond aggregate task success.
\label{sec:exp}

\subsection{Implementation Details}

\paragraph{Models, Benchmarks, and Baselines.}
In our experiments, we use Qwen3-VL-2B and Qwen3-VL-8B~\citep{yang2025qwen3}.
For each model, we first mid-train the backbone on mixed data from all three
domains to obtain a general student. This checkpoint initializes every
trainable student in our experiments. We then apply SFT separately on each
domain to obtain the three frozen domain-specific teachers. We test our method
on the GUI tasks of MobileWorld~\citep{kong2026mobileworld} and full set of OSWorld~\citep{xie2024osworld}.
We also evaluate a subset of WebVoyager~\citep{he2024webvoyager}, which contains 140 sample tasks. More information about this subset can be
found in Appendix~\ref{sec:webvoyager-subset}. The main setting consolidates all three
teachers in one run. The main comparison includes the domain-specific teachers, SFT, GRPO, Weight Soup~\citep{wortsman2022model}, TIES~\citep{yadav2023ties}, and
UI-MOPD~\citep{lian2026uimopd}. In the ablation study, we  remove the whole student side and its seperate rule. Besides, we also remove teacher side to demonstrate its effectiveness. And removing both student and teacher sides
of \csca recovers ordinary per-sample routed
OPD~\citep{lu2025onpolicydistillation}.

\paragraph{Training Details.}
Our training set contains 343k examples, including 93k from mobile, 50k from
desktop, and 200k from the web domain. Notably, the training data for GRPO, UI-MOPD, and \csca is constructed from the same SFT dataset by using the final step's action as the ground-truth label.  All runs are conducted on 32 H20 GPUs
and optimize the trainable LLM parameters with AdamW, using 1e-5 as learning rate, batch size 128. The vision parameters remain frozen. Each prompt
produces 4 rollouts. After each rollout, the rule-based
reward and parsed action spans determine the student-side allocation, while
the frozen domain-specific teacher scores the sampled tokens with a
training-only hint. During inference, we use vLLM~\citep{kwon2023efficient} as rollout engine. Further details are provided in
Appendix~\ref{sec:experimental-implementation}.

\paragraph{Evaluation Protocol.}
We evaluate fixed samples from each domain and use
Success Rate (SR) as the primary benchmark metric. Throughout the paper,
percentage differences denote absolute differences between percentage-valued
metrics rather than relative changes. Moreover, to measure
how much of each domain-specific teacher's capability is retained, we first
normalize a model merging method's SR by the corresponding teacher SR in each domain and then average these ratios. We define the resulting Teacher-Normalized Score
(TNS) for method $m$ as
\begin{equation}
\operatorname{TNS}(m)
=\frac{100}{|\mathcal D|}
\sum_{d\in\mathcal D}\frac{s_{m,d}}{s_{T_d,d}},
\label{eq:tns}
\end{equation}
where $s_{m,d}$ and $s_{T_d,d}$ are the SRs of method $m$ and the
domain-specific teacher on domain $d$, respectively. The higher the TNS is, the more capability of the domain-specific teachers is retained by the student. A TNS of $100\%$ matches
the domain-specific teachers on average.
Parameter-class breakdowns, parsing rules, and statistical tests are provided in
Appendix~\ref{sec:evaluation-details}.

\subsection{Main Results}

\paragraph{\csca outperforms the other baselines at both model scales.}
At 8B, \csca exceeds UI-MOPD on all three domains, improving mean SR by $2.0\%$
and TNS by $4.4\%$. It also leads Weight Soup and TIES on both metrics. At 2B,
although Weight Soup scores $2.1\%$ higher on
WebVoyager, \csca achieves the highest mean SR and TNS.

\paragraph{\csca is comparable to the teachers on average at 8B.}
Across all domains, \csca obtains a TNS of $99.9\%$. Its point
estimate is $3.4\%$ below the MobileWorld teacher but $2.4\%$ and
$2.1\%$ above the OSWorld and WebVoyager teachers, respectively. Its mean
SR is $0.4\%$ above the teacher mean.

\begin{table}[t]
\centering
\footnotesize
\setlength{\tabcolsep}{1.5pt}
\begin{tabularx}{\columnwidth}{@{}l*{3}{>{\centering\arraybackslash}X}@{}}
\toprule
Variant & MobileWorld  & OSWorld  & WebVoyager  \\
\midrule
\rowcolor{gray!12}
\csca & 34.2 & 45.3 & 74.3 \\
\ \ \ \ w/o student-side & 30.8 & 41.5 & 72.9 \\
\ \ \ \ \ \ \ \ w/o Rule 1 &31.6 & 45.0 &70.7\\
\ \ \ \ \ \ \ \ w/o Rule 2 &32.7 & 43.4 &74.3\\
\ \ \ \ \ \ \ \ w/o Rule 3 &33.3 & 43.9 &72.1\\
\ \ \ \ w/o teacher-side & 31.6 & 45.3 & 73.6 \\
\ \ \ \ w/o both sides & 29.1 & 44.4 & 70.0 \\
\bottomrule
\end{tabularx}
\caption{Ablation study of \csca on Qwen3-VL-8B. We report SR on MobileWorld, OSWorld, and
WebVoyager. }
\label{tab:csca_ablation}
\end{table}

\paragraph{\csca reduces the imbalance of weight merging.}
At 8B scale, Weight Soup exceeds the WebVoyager teacher by
$0.7\%$ but falls $8.5\%$ below the MobileWorld teacher.
\csca improves these margins to $+2.1\%$ and $-3.4\%$, respectively,
demonstrating more balanced cross-domain retention.

\suppressfloats[t]
\subsection{Ablation Study}

In Table~\ref{tab:csca_ablation}, student-side removal lowers SR by $3.4$, $3.8$, and $1.4$ points on MobileWorld,
OSWorld, and WebVoyager . Teacher-side hint
removal lowers MobileWorld and WebVoyager by $2.6$ and $0.7$ points, while OSWorld keeps
unchanged. Moreover, ablation results on separate student-side rules also demonstrate their effectiveness. 


\begin{table*}[t]
\centering
\footnotesize
\setlength{\tabcolsep}{2.0pt}
\begin{tabular*}{\textwidth}{@{\extracolsep{\fill}}lccccccccc@{}}
\toprule
& \multicolumn{3}{c}{Initial Student} & \multicolumn{3}{c}{Trained Student}
& \multicolumn{3}{c}{Change} \\
\cmidrule(lr){2-4}\cmidrule(lr){5-7}\cmidrule(lr){8-10}
Domain & Correct & Type$\times$ & Param.$\times$
       & Correct & Type$\times$ & Param.$\times$
       & $\Delta$Correct & $\Delta$Type$\times$
       & $\Delta$Param.$\times$ \\
\midrule
MobileWorld & 50.7 & 29.3 & 20.0 & 73.0 & 10.7 & 16.3
            & $+22.3$ & $-18.6$ & $-3.7$ \\
OSWorld     & 40.7 & 26.3 & 33.0 & 63.7 & 13.0 & 23.3
            & $+23.0$ & $-13.3$ & $-9.7$ \\
WebVoyager  & 35.3 & 37.0 & 27.7 & 53.0 & 20.0 & 27.0
            & $+17.7$ & $-17.0$ & $-0.7$ \\
\midrule
\textbf{Overall} & \textbf{42.2} & \textbf{30.9} & \textbf{26.9}
                 & \textbf{63.2} & \textbf{14.6} & \textbf{22.2}
                 & $\mathbf{+21.0}$ & $\mathbf{-16.3}$ & $\mathbf{-4.7}$ \\
\bottomrule
\end{tabular*}
\caption{Action outcomes of the initial and \csca-trained students, evaluated
on 300 held-out examples per domain. Correct denotes a correct action type and
all required parameters. Type$\times$ denotes responses without a correct
action type. Param.$\times$ denotes a correct action type with at least one
incorrect or missing required parameter.}
\label{tab:gate_coverage}
\end{table*}

\begin{figure*}[t]
\centering
\includegraphics[width=\textwidth]{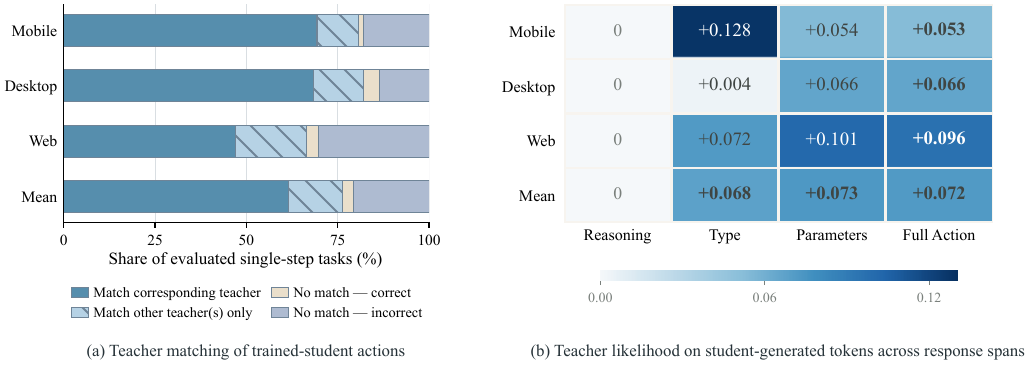}
\caption{Based on the trained student, we use 900 random samples (300 per domain) to generate response. (a) Each action falls into one of four mutually
exclusive groups: matching its corresponding domain-specific teacher, matching only other domain-specific
teacher, matching no teacher but is correct, or matching no teacher and
is incorrect. Mean values are averaged across domains.
(b) The heatmap compares how strongly each response matches its corresponding domain-specific teacher in the reasoning tokens action type tokens and action parameter tokens. Darker cells indicate a higher similarity to the teacher. All reported values are relative to the reasoning tokens, which serve as the zero-baseline for each domain.}
\label{fig:behavioral-inheritance}
\end{figure*}

\subsection{Do Student-Side Rules Work as Designed?}

\paragraph{All three student-side cases occur, and fully
correct actions increase by 21.0\%.}
In Table~\ref{tab:gate_coverage}, we compare the initial student with the
trained student on the same 900 held-out test examples, with 300 randomly
sampled examples from each domain. These examples are disjoint from the
training data.

Before training, $42.2\%$ of actions are fully correct, $30.9\%$ have a wrong action type and $26.9\%$ have the
correct type but an incorrect parameter. This indicates that all three
student-side allocation cases occur on the held-out test set, with a majority
of errors coming from the action type.

After training, the percentage of fully correct actions reaches $63.2\%$ ($+21.0\%$),
while parameter errors fall to $22.2\%$ ($-4.7\%$) and type errors fall
to $14.6\%$ ($-16.3\%$). The increase is consistent across domains, with
gains of $+22.3\%$ on MobileWorld, $+23.0\%$ on OSWorld, and $+17.7\%$ on
\mbox{WebVoyager}, which indicates that the student's performance improves across all domains.
Parameter-class results and task-level transitions from the initial to the
trained student are provided in Appendix~\ref{sec:allocation-profile-details}.

\subsection{Where Does Student's Behavior Come From?}

\paragraph{The trained student primarily matches the domain-specific teacher
and also produces a small set of successful novel actions.}
Based on 900 randomly sampled offline single-step tasks (300 per domain), we prompt the student and teachers to generate the corresponding answer action, and evaluate whether their action semantics align closely. Complete evaluation
rules are provided in Appendix~\ref{sec:teacher-matching-details}.

Figure~\ref{fig:behavioral-inheritance}(a) shows that 61.5\% of actions match
the domain-specific teacher. In comparison, 14.8\% match only other domain-specific teachers.
Thus, the student more often reproduces behavior consistent with its routed
domain-specific teacher than behavior found only in teachers from other
domain teachers.

Most student actions that do not align with any teacher are incorrect, accounting for $20.6\%$ of all evaluated tasks. However, $3.0\%$ of these unmatched actions are actually correct. These successful yet unaligned behaviors demonstrate that the student is capable of generating valid, novel actions.

\suppressfloats[t]
\begin{figure}[t]
\centering
\includegraphics[width=0.82\columnwidth]{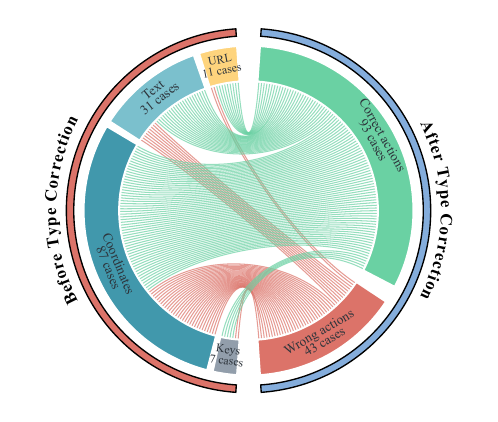}
\caption{For initially incorrect actions, correcting solely the action type allows the model to automatically recover most of the associated parameters. 
The student model regenerates all parameters from scratch, without access to the ground-truth reference parameters. 
Each row represents a single response.
The left section groups responses by the ground truth action type, while the right section indicates whether the regenerated action is correct.}
\label{fig:oracle-type-intervention}
\end{figure}

\subsection{On Which Tokens Do Student and Teacher Agree Most?}

\textbf{Action tokens generated by our trained student show stronger alignment with the
corresponding domain-specific teacher than reasoning tokens.}
Figure~\ref{fig:behavioral-inheritance}(b) compares the likelihood assigned by
teachers to reasoning and action tokens generated by our trained student.
The detailed calculation and statistical results are provided in
Appendix~\ref{sec:spanwise-likelihood-details}.

Compared with reasoning tokens, action tokens show an average likelihood gain
of $+0.072$. Specifically, the type and parameter tokens also show positive average gains
of $+0.068$ and $+0.073$ against reasoning tokens. 
This observation supports our action-aware approach to treat the action tokens as a distinct supervision target rather than weighting all response tokens uniformly.


\subsection{Can Action Failures Be Recovered from the Type Alone?}
\label{sec:oracle-type-intervention}

\paragraph{Correcting only the action type recovers a substantial percentage of wrong actions.}
From 900 sampled tasks across three domains, we isolate $136$ responses with incorrect action types. For these, we replace the predicted action type with the ground truth and prompt the frozen student to regenerate parameters using only its original context and reasoning (details in Appendix~\ref{sec:oracle-type-details}).

As shown in Figure~\ref{fig:oracle-type-intervention}, correcting only the action type recovers $68.4\%$ of the responses whose action type is wrong, while the remaining responses still fail due to wrong parameters. Specifically, $57$ out of $87$ actions requiring coordinate parameters are successfully recovered. This indicates that many wrong actions do not stem from a lack of capability in parameter generation. Instead, once the action type is corrected, the student model can often generate these parameters correctly. 
This finding also supports \csca's stronger supervision of the action type.

\subsection{Can \csca Correct Errors on  Samples with High-Disagreement?}
\label{sec:action-recovery}

\paragraph{\csca corrects weight-merge errors on both Click and Swipe cases.}
To evaluate Weight Soup and \csca, we construct a case study of 66 held-out MobileWorld tasks (29 \texttt{Click}, 37 \texttt{Swipe}). 
We filtered for spatial actions where all three teachers output valid coordinates with a maximum pairwise distance $>0.07$. 

Table~\ref{tab:action_recovery} reports the corresponding change in accuracy.
Weight Soup fails on 19 samples. Among them, \csca corrects 6 of the 13 Click errors and 4
of the 6 Swipe errors. One Click sample changes in the opposite direction,
from a correct prediction to an incorrect \csca prediction. Overall, \csca
corrects 10 Weight Soup errors while introducing one new error, increasing
accuracy from $47/66$ ($71.2\%$) to $56/66$ ($84.8\%$), a gain of $13.6\%$.

\subsection{Why Does the Student Not Consistently Surpass Its Teachers?}
\label{sec:student-teacher-gap}

\paragraph{Reasoning tokens dominate the response, thereby diluting the action supervision.}
As shown in Table~\ref{tab:trajectory-length}, we segment each response into reasoning and action spans. The action
accounts for only $3.9\%$--$7.1\%$ of response tokens, making it difficult for one student
to exceed every specialized teacher.

\begin{table}[t]
\centering
\small
\setlength{\tabcolsep}{3.5pt}
\begin{tabular*}{\columnwidth}{@{\extracolsep{\fill}}lccc@{}}
\toprule
Action
& \multicolumn{2}{c}{Correct actions}
& Change \\
\cmidrule(lr){2-3}
& Weight Soup & \csca & \\
\midrule
Click & 16/29 (55.2\%) & 21/29 (72.4\%) & +17.2\% \\
Swipe & 31/37 (83.8\%) & 35/37 (94.6\%) & +10.8\% \\
\bottomrule
\end{tabular*}
\caption{MobileWorld case study on high-disagreement samples. Model columns
report correct actions over the subset size, with accuracy in parentheses.
Change is \csca accuracy minus Weight Soup accuracy.}
\label{tab:action_recovery}
\end{table}

\begin{table}[t]
\centering
\footnotesize
\setlength{\tabcolsep}{1.8pt}
\begin{tabular*}{\columnwidth}{@{\extracolsep{\fill}}lccccc@{}}
\toprule
& \multicolumn{2}{c}{Token share (\%)} &
  \multicolumn{3}{c}{Mean interaction steps} \\
\cmidrule(lr){2-3}\cmidrule(lr){4-6}
Domain & Reasoning & Action & Successful & Failed & Gap \\
\midrule
MobileWorld & 95.9 & 4.1 & 19.2 & 29.4 & 10.2 \\
OSWorld     & 92.9 & 7.1 & 13.5 & 56.9 & 43.4 \\
WebVoyager  & 96.1 & 3.9 & 15.8 & 39.2 & 23.4 \\
\bottomrule
\end{tabular*}
\caption{Response-token composition and average interaction steps for the 8B
student. The step gap is the failed-trajectory mean minus the
successful-trajectory mean.}
\label{tab:trajectory-length}
\end{table}

\paragraph{Training on single-step data limits multi-step generalization.}
Table~\ref{tab:trajectory-length} shows that failed trajectories are longer
on average than successful ones, suggesting that long sequences are a
bottleneck.
We train \csca using only single-step supervision, but real-world GUI tasks require multi-step execution, misaligning the multi-step setting.
\section{Conclusion}
This work addresses how to consolidate multiple domain-specific GUI teachers
into a single agent while retaining their specialized capabilities. Our analysis shows that
weight merging degrades performance, particularly on high-disagreement
spatial samples, while ordinary routed OPD provides only limited training
signal to the short structured action. Moreover, the
short action span carries domain-specific behavior, which significantly affects the environment state. Therefore, we propose \csca to address this mismatch with
structured action signal re-allocation and a training-only teacher
hint, directing additional supervision according to action correctness and
structure. Across MobileWorld, OSWorld, and WebVoyager at two model scales,
\csca achieves the highest mean success rate among unified methods within
three domains and best preserves the capabilities of the domain-specific
teachers. Specifically, at the 8B scale, it improves mean SR by $2.0\%$ over
the strongest baseline and achieves almost the same average performance with teachers, suggesting that robust multi-platform
consolidation benefits from explicit supervision of structured actions.

\clearpage
\bibliography{refs}

\clearpage
\appendix
\section{Experimental Implementation Details}
\label{sec:experimental-implementation}

\subsection{Checkpoints and Data}
\label{sec:checkpoints-data}

\paragraph{Checkpoint construction.}
For each model scale, we mid-train Qwen3-VL on the mixed mobile,  desktop, and web data to obtain a shared general-student checkpoint. Every
trainable student in the main comparison, ablations, and analyses is
initialized from this checkpoint. We independently apply domain-specific SFT
to three copies of the same checkpoint to obtain the MobileWorld, OSWorld, and
WebVoyager teachers. These domain-specific teachers remain frozen during
distillation.

\paragraph{Training data.}
The data are subject to privacy requirements and
non-disclosure agreements and therefore cannot be released. The collection
differs from available public datasets in scale, domain composition, and
production conditions.
Our training corpus contains 343k proprietary industrial examples.
200k from the web domain, 93k from the MobileWorld domain, and 50k
from the OSWorld domain. 

\paragraph{WebVoyager subset.}
\label{sec:webvoyager-subset}
The fixed evaluation subset contains 140 tasks, with ten tasks from each of
14 websites. The task IDs are listed below.

\begingroup
\raggedright
\noindent\textbf{Allrecipes.}
\verb|Allrecipes--43|, \verb|Allrecipes--6|,
\verb|Allrecipes--11|, \verb|Allrecipes--17|,
\verb|Allrecipes--34|, \verb|Allrecipes--2|,
\verb|Allrecipes--14|, \verb|Allrecipes--36|,
\verb|Allrecipes--41|, \verb|Allrecipes--35|.
\par
\noindent\textbf{Amazon.}
\verb|Amazon--32|, \verb|Amazon--35|, \verb|Amazon--17|,
\verb|Amazon--33|, \verb|Amazon--8|, \verb|Amazon--28|,
\verb|Amazon--12|, \verb|Amazon--10|, \verb|Amazon--14|,
\verb|Amazon--37|.
\par
\noindent\textbf{Apple.}
\verb|Apple--13|, \verb|Apple--21|, \verb|Apple--33|,
\verb|Apple--10|, \verb|Apple--24|, \verb|Apple--38|,
\verb|Apple--19|, \verb|Apple--18|, \verb|Apple--26|,
\verb|Apple--34|.
\par
\noindent\textbf{ArXiv.}
\verb|ArXiv--7|, \verb|ArXiv--30|, \verb|ArXiv--37|,
\verb|ArXiv--2|, \verb|ArXiv--14|, \verb|ArXiv--20|,
\verb|ArXiv--35|, \verb|ArXiv--33|, \verb|ArXiv--17|,
\verb|ArXiv--23|.
\par
\noindent\textbf{BBC News.}
\verb|BBC News--40|, \verb|BBC News--3|,
\verb|BBC News--6|, \verb|BBC News--15|,
\verb|BBC News--35|, \verb|BBC News--26|,
\verb|BBC News--19|, \verb|BBC News--24|,
\verb|BBC News--36|, \verb|BBC News--17|.
\par
\noindent\textbf{Booking.}
\verb|Booking--15|, \verb|Booking--0|, \verb|Booking--29|,
\verb|Booking--8|, \verb|Booking--43|, \verb|Booking--35|,
\verb|Booking--19|, \verb|Booking--18|, \verb|Booking--22|,
\verb|Booking--33|.
\par
\noindent\textbf{Coursera.}
\verb|Coursera--21|, \verb|Coursera--29|,
\verb|Coursera--31|, \verb|Coursera--14|,
\verb|Coursera--16|, \verb|Coursera--37|,
\verb|Coursera--40|, \verb|Coursera--41|,
\verb|Coursera--23|, \verb|Coursera--18|.
\par
\noindent\textbf{ESPN.}
\verb|ESPN--34|, \verb|ESPN--21|, \verb|ESPN--38|,
\verb|ESPN--12|, \verb|ESPN--14|, \verb|ESPN--41|,
\verb|ESPN--26|, \verb|ESPN--8|, \verb|ESPN--37|,
\verb|ESPN--1|.
\par
\noindent\textbf{GitHub.}
\verb|GitHub--16|, \verb|GitHub--27|, \verb|GitHub--25|,
\verb|GitHub--5|, \verb|GitHub--7|, \verb|GitHub--18|,
\verb|GitHub--17|, \verb|GitHub--0|, \verb|GitHub--39|,
\verb|GitHub--10|.
\par
\noindent\textbf{Google Flights.}
\verb|Google Flights--6|, \verb|Google Flights--23|,
\verb|Google Flights--35|, \verb|Google Flights--8|,
\verb|Google Flights--15|, \verb|Google Flights--16|,
\verb|Google Flights--3|, \verb|Google Flights--20|,
\verb|Google Flights--22|, \verb|Google Flights--5|.
\par
\noindent\textbf{Google Map.}
\verb|Google Map--8|, \verb|Google Map--9|,
\verb|Google Map--5|, \verb|Google Map--19|,
\verb|Google Map--22|, \verb|Google Map--20|,
\verb|Google Map--17|, \verb|Google Map--39|,
\verb|Google Map--11|, \verb|Google Map--28|.
\par
\noindent\textbf{Google Search.}
\verb|Google Search--3|, \verb|Google Search--24|,
\verb|Google Search--21|, \verb|Google Search--35|,
\verb|Google Search--27|, \verb|Google Search--20|,
\verb|Google Search--41|, \verb|Google Search--9|,
\verb|Google Search--38|, \verb|Google Search--5|.
\par
\noindent\textbf{Huggingface.}
\verb|Huggingface--36|, \verb|Huggingface--33|,
\verb|Huggingface--17|, \verb|Huggingface--37|,
\verb|Huggingface--7|, \verb|Huggingface--4|,
\verb|Huggingface--5|, \verb|Huggingface--8|,
\verb|Huggingface--29|, \verb|Huggingface--12|.
\par
\noindent\textbf{Wolfram Alpha.}
\verb|Wolfram Alpha--36|, \verb|Wolfram Alpha--5|,
\verb|Wolfram Alpha--32|, \verb|Wolfram Alpha--8|,
\verb|Wolfram Alpha--1|, \verb|Wolfram Alpha--0|,
\verb|Wolfram Alpha--33|, \verb|Wolfram Alpha--17|,
\verb|Wolfram Alpha--29|, \verb|Wolfram Alpha--41|.
\par
\endgroup

\subsection{Training and Baseline Configurations}
\label{sec:training-baselines}

\paragraph{Compute and shared training setup.}
All training runs use 32 H20 GPUs for one epoch. We optimize the
trainable LLM parameters with AdamW using a global batch size of 128, zero
weight decay, and a maximum sequence length of 16,384 tokens. The learning
rate increases linearly during the first 3\% of training and then remains
constant at $1\times10^{-5}$. Each prompt produces four student rollouts.
Each training example contains one output action, although its input may
include interaction history. All distillation variants use the same student
initialization, routed domain data, and frozen domain-specific teachers. The
vision tower is bit-identical across teachers and remains frozen throughout
training. Besides, we separately disable the three rules in the student-side allocation to evaluate their effectiveness.

\paragraph{Baseline and ablation controls.}
The main comparison uses the complete method, including student-side
allocation and the teacher-side hint. The component ablation disables
either operation while keeping the remaining setup fixed. Disabling both
recovers ordinary routed OPD exactly.


\paragraph{Amplification coefficient.}
We set $\beta=1$ in all experiments, which doubles the weight of the selected
action tokens, and keep it fixed across model scales and domains without
domain-specific tuning.

\subsection{Hint and Allocation Implementation}
\label{sec:hint-allocation-implementation}

\paragraph{Training algorithm.}
Algorithm~\ref{alg:csca} summarizes the distributed implementation from
launch through actor updates. The worker count $P$ and resource placement are
read from the training configuration rather than hard-coded in the method.
The configured update count $K$ covers the single training epoch.
The Ray reward actor supplies only the exact-action acceptance gate; the actor
update uses the routed reverse-KL distillation advantage and does not optimize
a task-reward objective.

For sample $i$, let $a_i^*=(z_i^*,p_i^*)$ be the reference action and let
$y_i=(y_{i,1},\ldots,y_{i,|y_i|})$ be the response sampled from the student.
At token position $t$, $\ell_{i,t}^T$ is the log probability assigned to the
sampled token by the routed teacher using the type-hinted prompt, whereas
$\ell_{i,t}^S$ is the log probability assigned by the student using the
original prompt:
\[
\begin{aligned}
\ell_{i,t}^T
&=\log\pi_{T_{d_i}}(y_{i,t}\mid\mathcal P_T(x_i,z_i^*),y_{i,<t}),\\
\ell_{i,t}^S
&=\log\pi_\theta(y_{i,t}\mid\mathcal P_S(x_i),y_{i,<t}).
\end{aligned}
\]
The three binary masks identify the reasoning, action-type, and parameter
positions in the sampled response:
\[
\begin{aligned}
m_{i,t}^{r}&=\mathbf{1}[t\in\mathcal I_{\mathrm{reason}}],\\
m_{i,t}^{z}&=\mathbf{1}[t\in\mathcal I_{\mathrm{type}}],\\
m_{i,t}^{p}&=\mathbf{1}[t\in\mathcal I_{\mathrm{param}}].
\end{aligned}
\]
These disjoint masks partition the response tokens. Parameter tokens include
coordinates, text, URLs, keys, and other values required by the predicted
type-specific schema. We use $\hat z_i=\textsc{ParseType}(y_i)$ for the type
parsed from the student response. Brackets denote a binary indicator, so
$[\hat z_i\ne\textsc{Fail}\land\hat z_i=z_i^*]$ equals one exactly when parsing
succeeds and the predicted type matches the reference type, and equals zero
otherwise. 

\begin{algorithm*}[t]
\caption{ \csca Training Algorithm}
\label{alg:csca}
\footnotesize
\begin{algorithmic}[1]
\REQUIRE Config $\mathcal C$; routed data $\mathcal D$; student $\pi_\theta$;
teachers $\{T_d\}$; coefficient $\beta$
\STATE \textbf{Stage 0: launch}\quad Read $P$ and $K$ from $\mathcal C$; set
$k\gets0$; use \textsc{Torchrun} to start $P$ processes; initialize Ray and
one runner.
\STATE \textbf{Stage 1: initialize}\quad Load the tokenizer/processor,
dataloader, reward actors, and shared GPU resource pool.
\FORALL{$p\in\{1,\ldots,P\}$ \textbf{in parallel}}
    \STATE Load the sharded actor, vLLM rollout engine, and frozen teachers
    into one FSDP worker.
\ENDFOR
\STATE \textbf{Stage 2: train}
\WHILE{$k<K$}
    \STATE $X\gets\textsc{NextBatch}(\mathcal D)$;\quad
    $B\gets\textsc{ExpandEachPrompt}(X,4)$.
    \STATE $y_i\sim\pi_\theta(\cdot\mid x_i)$ for each rollout slot $i\in B$.
    \STATE Asynchronously compute
    $c_i\gets\textsc{ExactAccept}(y_i,a_i^*)$ for each rollout.
    \FORALL{$i\in B$ \textbf{in parallel}}
        \STATE Route source $d_i$ to $T_{d_i}$; set
        $h_i^T=(x_i,\textsc{TypeHint}(z_i^*))$.
        \STATE Score $y_i$ under $T_{d_i}$ with $h_i^T$ and under
        $\pi_\theta$ with $x_i$ to obtain $\ell_{i,t}^T$ and $\ell_{i,t}^S$.
        \STATE Build $m_{i,t}^{r},m_{i,t}^{z},m_{i,t}^{p}$ from the sampled
        token IDs.
        \STATE $\hat z_i\gets\textsc{ParseType}(y_i)$;\quad
        $q_i\gets[\hat z_i\ne\textsc{Fail}\land\hat z_i=z_i^*]$.
        \STATE $\widehat A_{i,t}^{H}\gets
        \operatorname{sg}[\ell_{i,t}^T-\ell_{i,t}^S]$.
        \IF{$c_i$}
            \STATE $w_{i,t}\gets0$\hfill\COMMENT{fully accepted}
        \ELSIF{$q_i$}
            \STATE $w_{i,t}\gets m_{i,t}^{r}
            +(1+\beta)(m_{i,t}^{z}+m_{i,t}^{p})$
            \hfill\COMMENT{amplify action}
        \ELSE
            \STATE $w_{i,t}\gets m_{i,t}^{r}+(1+\beta)m_{i,t}^{z}$
            \hfill\COMMENT{amplify type; mask params}
        \ENDIF
        \STATE $\widehat A_{i,t}^{\csca}\gets w_{i,t}\widehat A_{i,t}^{H}$.
    \ENDFOR
    \STATE $\mathcal L\gets-\frac{1}{|B|}\sum_{i\in B}
    \frac{1}{|y_i|}\sum_t\widehat A_{i,t}^{\csca}
    \log\pi_\theta(y_{i,t}\mid x_i,y_{i,<t})$.
    \STATE Update
    \STATE $k\gets k+1$.
\ENDWHILE
\STATE \textbf{Stage 3: finalize}\quad Save the actor, TensorBoard events, and
per-token distillation logs.
\end{algorithmic}
\end{algorithm*}

\paragraph{Teacher-side hint.}
For each training sample, we construct a text hint from only the ground-truth
action using the template \texttt{Hint: the correct action for this step
is a \{HINT\} action.} We append the hint to the selected domain-specific
teacher's user turn, after the original instruction and before the assistant
response. The teacher then scores exactly the
student-generated response tokens. Disabling the hint restores the original
teacher prompt.

\paragraph{Mask construction.}
We preserve ordinary routed OPD trajectory collection and build the allocation
masks after rollout.
Incremental cumulative decoding locates the reasoning, type, parameter, and
action spans without re-encoding the response or invoking another model.
The action-acceptance component of the existing rollout reward determines
whether the sampled action is fully correct. The selected domain-specific teacher then
scores the student tokens without decoding. Under an incorrect type, parameter
tokens are masked, while only the type span receives additional weight and
reasoning retains its ordinary distillation advantage.

\subsection{Exact Action Evaluator}
\label{sec:exact-action-evaluator}

\paragraph{Exact action acceptance.}
\label{sec:full-response-acceptance}
The correctness gate reuses the action-acceptance component of the existing
rule-based training reward. It is an offline evaluator signal, not task success
from live interaction. A sampled action is fully correct only when the evaluator
accepts its type and every parameter required by the reference type-specific
schema. Partial parameter credit does not pass the gate. The separate response-
format reward is unchanged by our method but does not determine student-side
allocation.

Both the predicted and reference actions are parsed into an action type and
its type-specific parameters. A parse failure or type mismatch fails the
action-acceptance gate. An unparseable action follows the wrong-type branch that
the implementation retains ordinary reasoning supervision, strengthens an
identifiable type token when available, and masks tokens assigned to the
parameter span. Once the type matches, the evaluator applies the
action-specific parameter rule in Table~\ref{tab:action_acceptance_rules}.

\begin{table}[t]
\centering
\footnotesize
\setlength{\tabcolsep}{3pt}
\begin{tabular}{@{}p{0.31\columnwidth}p{0.62\columnwidth}@{}}
\toprule
Action family & Full-acceptance rule \\
\midrule
\texttt{Click}, \texttt{LongPress}, \texttt{Hover}, \texttt{DoubleClick}
& Point distance is at most $0.07$ times the shorter screen side. \\
\texttt{Drag}
& Both spatial endpoints are at most $0.14$ times the shorter screen side from their references. \\
MW/OSW \texttt{Scroll} (direction)
& Direction matches exactly and both endpoints are at most $0.07$ from their references. \\
MW/OSW \texttt{Scroll} (no direction)
& Both endpoints are at most $0.14$ from their references. \\
WV \texttt{Scroll}
& Direction matches exactly. \\
Text/app/URL
& Exact match or token F1 above $0.5$. \\
Hotkeys
& Exact match after stripping surrounding whitespace. \\
Empty schema
& Action type matches. \\
\bottomrule
\end{tabular}
\caption{Correctness rules for each type-specific parameter.
Partial credit does not pass the binary allocation gate.}
\label{tab:action_acceptance_rules}
\end{table}

\subsection{Analysis Protocols}
\label{sec:analysis-protocols}

\paragraph{Allocation-profile experiment.}
The allocation profile compares the Qwen3-VL-2B student checkpoint
with its corresponding \csca-trained checkpoint. We sample 300 held-out test
examples per domain. These examples are disjoint from the
training data. Prompts longer than 14k tokens are
excluded rather than truncated. Both checkpoints use the same prompt template
and rollout settings, with temperature $0.9$, top-$p$ $1.0$, no top-$k$
truncation, and a maximum completion length of 1024 tokens. All 900 task IDs
are matched between checkpoints for the transition analysis.

\paragraph{Teacher-matching analysis.}
\label{sec:teacher-matching-details}
We compare the trained student's parsed final action with the
actions produced by all three frozen domain-specific teachers on the same
offline single-step task. The matched set contains 900 tasks, with 300 from
each domain. An action match requires
the same action type and accepted type-specific parameters. Coordinate-valued
parameters use the $0.07$ normalized-distance threshold. A shared match means
that the student matches at least two teachers. For the main figure, all
actions matching the domain-specific teacher are merged into one category,
whether or not they also match another domain-specific teacher. Actions that
do not match the domain-specific teacher are divided into matches with other
domain-specific teachers and matches with no teacher. 
No-teacher matches are divided by whether the existing evaluator accepts the
student action for the reference step.

\paragraph{Spanwise teacher-likelihood analysis.}
\label{sec:spanwise-likelihood-details}
 All
three frozen teachers score every response under teacher forcing, without
decoding a new response. Let $M_{d,t}^{(s)}$ denote the mean length-normalized
log likelihood assigned by teacher $t$ to span $s$ of responses from domain
$d$. Because raw scores contain both response-domain and teacher-wide
calibration effects, we compute the two-way-centered interaction
\begin{equation}
R_{d,t}^{(s)}=M_{d,t}^{(s)}-\bar M_{d,\cdot}^{(s)}
-\bar M_{\cdot,t}^{(s)}+\bar M_{\cdot,\cdot}^{(s)}.
\end{equation}
For each domain, the reported value for span $s$ is
$E_d^{(s)}=R_{d,d}^{(s)}-R_{d,d}^{(\mathrm{reasoning})}$. Reasoning is
therefore zero by construction, and positive values indicate a larger
domain--teacher interaction than on reasoning tokens. The type and parameter
columns are diagnostic subspans of the full action and are not additive. We
compute uncertainty with 2,000 domain-stratified bootstrap resamples over task
IDs and macro-average only after computing the statistic within each domain.

\paragraph{Disagreement-conditioned diagnostics.}
We first select samples with a reference spatial action and parseable
coordinates from all three frozen teachers. High disagreement means that the
maximum pairwise teacher distance exceeds the single-step acceptance radius
($0.07$). 

\paragraph{Oracle type intervention.}
\label{sec:oracle-type-details}
We greedily decode $900$ randomly sampled responses. We retain responses whose predicted type is wrong and parseable,
whose reference action requires parameters, and whose type span can be located
reliably. This gives $136$ responses: $36$ from MobileWorld, $52$ from OSWorld,
and $48$ from WebVoyager.

For each response, we keep the image, interaction history, and generated
reasoning fixed, replace only the predicted type with the reference type, and
let the same frozen student regenerate the parameters. The student receives no
teacher output or reference parameters. Restoring the original type reproduces
all $136$ original responses, confirming that the intervention changes only
the type.

\section{Evaluation Metrics and Statistical Tests}
\label{sec:evaluation-details}

\paragraph{Structured-action diagnostics.}
The primary benchmark metric is task-level SR. To locate action errors,
we additionally report action-type accuracy and parameter acceptance conditional on a correct type.
Conditional parameter results are grouped by the reference parameter
class and include the
denominator of every class. These metrics diagnose the source of task failures
but do not replace SR.

For the disagreement-conditioned spatial-action analysis, we report
single-step success rate. Each sample contains one reference spatial action. A
prediction is a single-step success only when it has the correct action type, a
parseable coordinate, and $d_\text{norm}\leq0.07$. Wrong-type and unparseable
predictions are failures. We compute this rate over the fixed
spatial-action subset and report its denominator and success count. It is
distinct from benchmark task-level SR. 

\section{Action-Span Token and Signal Allocation}
\label{sec:action-span-dilution}

We draw $300$ responses from each domain and locate the final structured action
using the same token boundaries used by the training masks. For each response,
we divide the number of action tokens by the number of valid response tokens.
The action accounts for $4.1\%$ of MobileWorld tokens, $7.1\%$ of OSWorld
tokens, and $3.9\%$ of WebVoyager tokens.

The archived training logs retain token-level distillation values by
normalized response position. Across domains, action tokens receives $4.0\%$--$7.0\%$ of the total
logged distillation signal (Table~\ref{tab:action_span_dilution}), compared with $13\%$--$14\%$ for the first ten
response tokens.

\begin{table}[t]
\centering
\footnotesize
\setlength{\tabcolsep}{3.5pt}
\begin{tabular*}{\columnwidth}{@{\extracolsep{\fill}}llr@{}}
\toprule
Measure & Domain or response region & Share (\%) \\
\midrule
Action tokens & MobileWorld ($N=300$) & 4.1 \\
Action tokens & OSWorld ($N=300$) & 7.1 \\
Action tokens & WebVoyager ($N=300$) & 3.9 \\
\midrule
Distillation signal & Action Tokens & 4.0--7.0 \\
\bottomrule
\end{tabular*}
\caption{Exact action-token share and distillation signal share.}
\label{tab:action_span_dilution}
\end{table}

\section{Allocation Profile Breakdown}
\label{sec:allocation-profile-details}


Table~\ref{tab:allocation_param_class} breaks the profile down by the reference
parameter schema. Coordinates dominate the sample; the URL subset has only
five examples and supports no class-specific conclusion. These classes cover
841 examples, while 59 empty-schema actions remain in the aggregate results in
Table~\ref{tab:gate_coverage}. Empty-schema actions require no parameters.
 A type match is Correct, whereas a wrong is
Type$\times$. Param.$\times$ cannot occur. Of these 59 actions, the initial
student has 18 Correct and 41 Type$\times$, while the trained student has
44 Correct and 15 Type$\times$.

\begin{table}[!t]
\centering
\scriptsize
\setlength{\tabcolsep}{2.0pt}
\begin{tabular*}{\columnwidth}{@{\extracolsep{\fill}}lrrrrrrr@{}}
\toprule
& $N$ & \multicolumn{3}{c}{Initial} & \multicolumn{3}{c}{Trained} \\
\cmidrule(lr){3-5}\cmidrule(lr){6-8}
Class & & C & T$\times$ & P$\times$ & C & T$\times$ & P$\times$ \\
\midrule
Coordinates & 678 & 45.1 & 25.1 & 29.8 & 62.4 & 14.3 & 23.3 \\
Text        & 121 & 36.4 & 38.0 & 25.6 & 60.3 &  5.0 & 34.7 \\
URLs        &   5 &  0.0 &100.0 &  0.0 & 40.0 & 60.0 &  0.0 \\
Keys        &  37 & 32.4 & 43.3 & 24.3 & 73.0 & 27.0 &  0.0 \\
\bottomrule
\end{tabular*}
\caption{Allocation profile by reference parameter class. C, T$\times$, and
P$\times$ denote correct, wrong-type, and wrong-parameter outcomes.}
\label{tab:allocation_param_class}
\end{table}

\end{document}